\documentclass[fleqn,10pt]{wlscirep}
\usepackage[utf8]{inputenc}
\usepackage[T1]{fontenc}
\usepackage{amsmath,amssymb}
\usepackage{multirow}
\usepackage{graphicx}
\usepackage{array}
\usepackage{caption, subcaption}
\def\ensemble{COURAGE}
\def\covid{COVID-19}

\usepackage{soul}

\title{\textbf{COU}nty agg\textbf{R}egation mixup \textbf{A}u\textbf{G}m\textbf{E}ntation (COURAGE) {COVID-19} Prediction}

\author[1]{Siawpeng Er}
\author[2,*]{Shihao Yang}
\author[3,*]{Tuo Zhao}
\affil[1,2,3]{ H. Milton Stewart School of Industrial and Systems Engineering\\Georgia Institute of Technology\\Atlanta, GA 30332, USA}
\affil[*]{To whom correspondence should be addressed: shihao.yang@isye.gatech.edu, tourzhao@gatech.edu}

\begin{abstract}
The global spread of {\covid}, the disease caused by the novel coronavirus SARS-CoV-2, has casted a significant threat to mankind. As the {\covid} situation continues to evolve, predicting localized disease severity is crucial for advanced resource allocation. This paper proposes a method named COURAGE (\underline{COU}nty agg\underline{R}egation mixup \underline{A}u\underline{G}m\underline{E}ntation) to generate a short-term prediction of 2-week-ahead {\covid} related deaths for each county in the United States, leveraging modern deep learning techniques. Specifically, our method adopts a self-attention model from Natural Language Processing, known as the transformer model, to capture both short-term and long-term dependencies within the time series while enjoying computational efficiency. Our model solely utilizes 
publicly available information for {\covid} related confirmed cases, deaths, community mobility trends and demographic information, and can produce state-level predictions as an aggregation of the corresponding county-level predictions. Our numerical experiments demonstrate that our model achieves the state-of-the-art performance among the publicly available benchmark models.
\end{abstract}
\begin{document}

\flushbottom
\maketitle

\section{Introduction}
{\covid} has been spreading globally and affected almost every country since 2020. In the United States (US), the {\covid} pandemic started spreading in January 2020, and in March, the daily number of confirmed cases and number of deaths rose to an alarming stage\cite{coviddatatracker}. To control the rapid spread of {\covid}, policymakers in many states imposed movement restrictions and partial confinement to everyone. For example, companies, educational institutes and public places were forced to close or work in remote settings. This has significantly affected the nationwide economy \cite{covideconomy} and posed challenges to our daily life. More importantly, the severity of {\covid} resulted in the loss of lives and might have caused serious long-term complications \cite{covidlongterm}. With nearly $31$ million cases and $0.56$ million deaths \cite{newyorktimeweb} in the US alone, {\covid} has become a serious threat to mankind.  

Since the beginning of the pandemic, different parties \cite{nyt, pmid32087114, covidtrackingproject, mobility} have made concerted efforts to collecting and publishing {\covid} related data, including confirmed cases, deaths, hospitalization information, demographics\textcolor{blue}{,} and community mobility. Based on the publicly available data, researchers have built predictive models to study the disease dynamics. These efforts include compartmental models such as variants of Susceptible-Infectious-Recovered (SIR) models \cite{harvardsimulation, googlecloud}, statistical models using regression \cite{altieri2020Curating} or time series analysis \cite{Lampos2021}. Besides, researchers have also applied agent-based simulation modeling \cite{Kerr2020, Germann2020} and deep learning models \cite{Rodr2020, stan2021, jin2020interseries} for predicting {\covid} dynamics. Moreover, the Centers for Disease Control and Prevention (CDC) has been leading a collaborative effort to produce an ensemble model from different research groups \cite{Ray2020.08.19.20177493} (See more detailed discussions of the related work in {\covid} dynamics prediction in a later section).

Two key measurements used by various research groups in the study of the spread of {\covid} are the number of confirmed cases and the number of deaths. Both measurements serve to measure the disease dynamics of {\covid}. It should be noted that both the number of confirmed cases and the number of deaths are subject to similar biases that may affect the accuracy of the data, primarily due to reporting criteria \cite{covidtrackingproject} or administration delay \cite{newyorkundercount}. However, the number of confirmed cases is often subject to additional measurement error due to the number of people tested and the testing procedure. For example, some datasets \cite{covidtrackingproject} only present confirmed cases as the number of people having positive results for a completed polymerase chain reaction test. Such coarse-grained testing numbers may undercount the true spread of the {\covid}, while introducing a significant bias to predictive models. Despite the caveat, the number of deaths is still a relatively better indicator of the intensity of {\covid} for policymakers to make decisions. As such, we will focus on predicting {\covid}-related deaths in this study.


Furthermore, due to different disease spread severity in different geographical areas, policymakers need to tailor policies based on the particular local situation for the state or the county. Hence, accurate prediction at the state and county level is crucial for informed decisions. 
Indeed, the county-level analysis could provide insightful information at finer granularity for policymakers \cite{Li2020.12.02.20234989, stan2021,zhu2020highresolution,Chande2020}. Policymakers can maximize resource allocation efficiency and react promptly in the legislative areas that require urgent attention. Therefore, we aim to build a model which can make short-term predictions for the number of deaths at the county level. We can also produce predictions for the number of deaths at the state level by a simple aggregation of our county-level predictions. Such a model could help both the state and local governments to make an informed decision based on the predictions at the corresponding state and county levels. 

The short-term prediction of {\covid} dynamics is essentially a classical time series modeling problem. For each day, we collect {\covid} related data as the input, and the desired output would be the predicted number of deaths for the next 2 weeks. On the next day, we obtain the new data from one additional day, and update our predictions for the next 2 weeks starting from the new date. This paper develops a fully data-driven approach to model {\covid} dynamics. Specifically, we build a self-attention deep learning model, which takes the input data from multiple sources and predicts the number of {\covid}-related deaths for the future 2 weeks at the county level in the United States. The state-level prediction is then obtained by summing up all county-level predictions of the corresponding state. Moreover, we propose a sequence mix-up augmentation approach to further improve the training of our transformer model \cite{zhang2018mixup}. We carefully evaluate our proposed methods under different training for different periods. We further compare our best model with other benchmark models to show its strength and usability. 

Unlike most existing deep learning approaches based on recurrent neural networks (RNN) such as LSTM-RNN \cite{HochSchm97}, our proposed approach is based on a current state-of-art self-attention model in Natural Language Processing, also known as the transformer model, which is able to capture both short-term and long-term dependencies within the time series and enjoys computational efficiency. In particular, the basic building blocks of the transformer model are the self-attention modules, which directly model dependencies on previous time steps by assigning attention scores. A large score between two events implies a strong dependency, while a small score implies a weak one. In this way, the modules are able to adaptively select time steps that are at any temporal distance from the current time step. Therefore, the transformer model has the ability to capture both short-term and long-term dependencies. Moreover, the non-recurrent structure of the transformer facilitates efficient training of multi-layer models. Practical transformer models can be as deep as dozens of layers, where deeper layers capture higher order dependencies. The ability to capture such dependencies creates models that are more powerful than RNNs, which are often shallow. Also, the transformer models allow full parallelism when calculating dependencies across all time steps, i.e., the computation between any two time steps is independent of each other. This yields a model presenting strong computational efficiency.



\subsection{Related Work}
Epidemic prediction is a time series prediction problem. Given a sequence of time with corresponding data, a model needs to predict the target incidence in the future time. There are four main classes of predictive models for epidemic prediction: the compartmental model,  simulation modeling, statistical model, and deep learning model. Moreover, for the same class of models, the final prediction may be an ensemble of several other models. In fact, on the CDC website, the final CDC prediction is obtained by ensembling all the submitted  models \cite{Ray2020.08.19.20177493}.

\noindent$\bullet$ \textbf{Compartmental model} is one of the most widely used model types for modeling epidemic diseases. It characterizes the disease spread dynamics using systems of ordinary differential equations. One of the most successful compartmental models is the SIR model, which is used to predict disease progression within the population in one area. In the SIR \cite{Harko_2014, Chen_2020} model, the population is assigned to Susceptible (S), Infectious (I), or Recovered (R) mode. One variant of SIR model is SEIR model \cite{Hethcote00themathematics, JTD36385, Xu2020, Guo2020.06.18.20134916}, which introduces additionally Exposed (E) mode. Other variants of the SIR model include the SIRD \cite{Caccavo2020} model, with additional Deceased (D) mode.  Any transitions from one mode to another mode (i.e., the disease spreading dynamics) are modeled as differential equations, often in the form of a transition matrix. Compartmental models are hard to be widely used due to the difficulties to determine the hyperparameters in every differential equation used \cite{baek2021limits}. One highly successful compartmental model for {\covid} predictions is from Karlen's group \cite{karlen2020characterizing}. Karlen's model uses discrete-time difference instead of ordinary differential equations to model the transition matrix.

\noindent$\bullet$ \textbf{Simulation modeling} uses computer simulation to model different components in the studied environment and observe their interactions. Cellular automata and agent-based simulation are two simulation modeling techniques used to model complex systems \cite{sayama2015}. In {\covid} prediction, several groups \cite{Kerr2020,Germann2020} use agent-based simulation due to its flexibility to simulate dynamic behaviours of systems with large number of entities. While highly flexible, agent-based simulation requires access to extensive computational resources. Besides, multiple simulations are needed for statistically sound observations, resulting in longer inference time. This limitation is noticeable when the simulated systems involve a large number of individual entities. 

\noindent$\bullet$ \textbf{Conventional statistical models} use regression methods to fit the data directly. Such models include ARIMA, Gaussian process regression, and linear regression, which are more flexible than compartmental models. However, most of the time, statistical model usage is limited by the need for more sophisticated hand-crafted features, which often requires knowledge from the domain experts.
For example, CLEP model \cite{altieri2020Curating} uses an ensemble model of an exponential predictor and a linear predictor. Model from Zhu et al. \cite{zhu2020highresolution} uses spatio-temporal information among counties to make predictions. Model from Lampos et al. \cite{Lampos2021} applies statistical model (Autoregressive model and Gaussian Process regression) to predict the disease's dynamics across countries in Europe and the United Kingdom based on online search data.

\noindent$\bullet$ \textbf{Deep Learning models} are deep neural networks that learn directly from Input data. Such models are more flexible than both compartmental models and conventional statistical models. Due to their representation capability, such models need a less sophisticated handcrafting preprocessing of the input data. In time series prediction problems, some common deep learning models include Long short-term memory (LSTM) \cite{HochSchm97}, Gated Recurrent Unit (GRU) \cite{cho2014learning}, and transformer \cite{vaswani2017attention, zuo2021transformer}. All the above models can capture intrinsic information from sequential data for accurate prediction. One limitation of such models is the need for large training data.  Concurrently with our work, there are other deep learning models including models from Ref.\citenum{stan2021} and Ref.\citenum{jin2020interseries} that utilize attention mechanism from transformer architecture.

Different models have their merits in their performance for different date ranges. During the beginning of the {\covid} outbreak, due to the limitations of the available data, we see predictions from the compartmental models or statistical models. With more data available, deep learning models are showing their advantages in the model flexibility and prediction accuracy. It is also more challenging for a model to make accurate predictions as the prediction granularity becomes finer. Intuitively, errors are more likely to be accumulated if a model is tasked to predict more targets than only a few targets. However, finer prediction granularity, such as prediction at the county level, is highly desired since local predictions help policymakers make tailored policies based on individual counties.  

\subsection{Our contribution}


In this paper, our goal is to predict the weekly total number of deaths at both county level and state level for the next two weeks, given the current week data. Each single-day data include the number of confirmed cases, the number of deaths, community mobility, and population. Our novelty is to connect established Natural Language Processing techniques with the COVID-19 time series prediction. In particular, we build a self-attention model, also known as the transformer model in Natural Language Processing, that is able to capture both the short and long term dependencies within the time series input data. 
Our model is build upon two main ideas. First, by aggregating the accurate predictions from county level to state level, our model shows strong performance for the state-level prediction task in all prediction periods. Then, we further improve our model, by experimenting with the feasibility of using data augmentation method. We implement mixup \cite{zhang2018mixup} as our data augmentation method at the input layer. To the best of our knowledge, this is the first application of mixup data augmentation in {\covid} data. Data augmentation further improves our model performance, and is particularly helpful when new trends emerge. Using these two core ideas, our proposed method, named COURAGE (\underline{COU}nty agg\underline{R}egation mixup \underline{A}u\underline{G}m\underline{E}ntation), is able to generate a short-term prediction of 2-week-ahead {\covid} related deaths for each county and state in the United States. When compared with other benchmark models, COURAGE shows strong performance across different periods, showing its strength and usability for the prediction of the {\covid} related number of deaths.

\section{Results}
\label{sec:evaluations}
To evaluate our proposed method, we compare county-level and state-level predictions using both the county-level and state-level testing sets with mean absolute error (MAE) as our comparison metrics. For county-level predictions, we compare the predictions from our {\ensemble} model with its two member models --- the County model and the Mixup model. We also compare our model with the baseline Naive model, which simply uses the previous week's reported total number of deaths as the prediction. In the state-level predictions, besides the previous mentioned models, we have an additional baseline State model, which is COURAGE prediction generated solely on the state-level input data without any county-level information or the mixup data augmentation. We compare the predictions according to the training period used, corresponding to $0.5$, $0.6$, $0.7$, and $0.8$ of the total dataset. We also present the performance of each model across multiple non-overlapping periods. Finally, we compare our models with the available models contributed to the CDC forecast website.


\subsection{Comparison Among Models}
We summarize our comparison among our models in Table \ref{tab:county_level_prediction}. We emphasize that our {\ensemble} model is an ensemble model that consists of two separate models: the County and the Mixup model. Those two can also serve as standalone models. We have the Naive model (a.k.a. persistence forecast model, which uses the previous week's reported number of deaths as the forecast) as a baseline model in the county-level prediction task. We use both the Naive model and the State model (prediction using only state-level inputs) as baseline models for the state-level prediction task.  

The County model and the Mixup model produce more accurate predictions than the Naive model in the county-level prediction task. By combining these two models, we improve our {\ensemble} model's prediction accuracy. 
We also get accurate predictions from the County model and the Mixup model in the state-level prediction task. They outperform both the Naive model and the State model. When combining both models, {\ensemble} has the best accuracy under different prediction periods, shy from the Week 2 predictions when using the training dataset dated from March 7, 2020, to December 1, 2020. We observe that the County model and the Mixup model have their strengths for different periods. Our {\ensemble} model often obtains the best of both models and produces the most accurate predictions.

\begin{table}[!ht]
    \centering
    \begin{tabular}{p{5cm}p{2.1cm}>{\raggedleft}p{2cm}>{\raggedleft}p{2cm}>{\raggedleft}p{2cm}>{\raggedleft\arraybackslash}p{2cm}}
    \hline
    \multirow{2}{*} & & \multicolumn{2}{c}{County Level}& \multicolumn{2}{c}{State Level}\\
    \cline{3-6}
    Training size & Method & Week 1 & Week 2  & Week 1 & Week 2\\
    \hline
    \multirow{4}{*}{0.5}
         & Naive & 2.0000 & 2.2765 & 51.5932 & 71.8391\\ 
        & State & - & - & 64.1419 & 81.9421 \\
    \multirow{2}{*}{(2020-03-07 to 2020-08-22)}
        & County & 1.9205 & 2.2808 & 48.0625 & 67.1583\\
        & Mixup  & 1.9227 & 2.2442 & 51.4493 & 63.4738\\
        & \ensemble & 1.8691 & 2.1602 & 47.7390 & 60.8701 \\
    \hline
    \multirow{4}{*}{0.6}
         & Naive & 2.2302 & 2.5698 & 58.7599 & 84.1373\\
         & State &- & - & 64.1419 & 81.9421 \\
    \multirow{2}{*}{(2020-03-07 to 2020-09-24)}
        & County & 2.1805 & 2.5331 & 53.0127 & 73.8226\\
        & Mixup  & 2.0932 & 2.3526 & 52.5268 & 70.6528\\
        & \ensemble & 2.0984 & 2.3633 & 51.1680 & 68.7564\\
    \hline
    \multirow{4}{*}{0.7}
         & Naive & 2.6275 & 3.0295 & 71.8298 & 102.1820\\ 
      &  State &  - & - & 101.2264 &  123.0914\\
     \multirow{2}{*}{(2020-03-07 to 2020-10-28)}
     &   County & 2.4912 & 2.8654  & 65.3611 & 85.1719 \\
     &   Mixup  & 2.5054 & 2.8034  & 66.9685 & 83.6469 \\
     & \ensemble & 2.4583 & 2.7547 & 64.9884 & 81.9787 \\
    \hline
    \multirow{4}{*}{0.8}
     & Naive &  3.0036 & 3.3499 & 80.5957 &  106.7175\\ 
     &  State & - & - & 118.9543 & 147.5468 \\
    \multirow{2}{*}{(2020-03-07 to 2020-12-01)}
     &   County & 2.8240 & 3.1477 & 74.7763 & 98.2125 \\
     &  Mixup  & 2.8258 & 3.0506 & 71.6854 & 84.4188 \\
     & \ensemble &  2.7670 & 2.9753 & 70.8481 & 86.3737\\
    \hline
    \end{tabular}
    \caption{Prediction of county-level and state-level weekly total number of deaths. The comparison metrics used is MAE, and the testing dataset starts from the 2020-08-23, 2020-09-25, 2020-10-29 and 2020-12-02 to 2021-02-07 for each corresponding training dataset.}
    \label{tab:county_level_prediction}
\end{table}

\subsection{Predictions for Different Periods}
We show the number of deaths prediction for different periods in Table \ref{tab:different_prediction_period}. Since we use different amount of datapoints as our training set, our training data consists of information across different periods. 
We take the non-overlapping period from each testing set as a separate out of the sample prediction period. In all the periods, the County and Mixup models produce better predictions than the Naive model. The result shows the feasibility of both models. For the Mixup model, its strength is visible in the last two periods of prediction. In the last prediction period, we use our trained model (using training set with data from 2020-03-07 to 2020-12-01) to predict the latest data with a prediction period from 2021-01-18 to 2021-03-14. Our model is able to predict accurately for the new data, as illustrated by plot of prediction for New York in Figure \ref{fig:timeline} and additional plots for other major states (Illinois, California, Texas, Arizona) in Supplementary Information. We mark in Figure \ref{fig:timeline} with light grey lines for different prediction date ranges as that of Table \ref{tab:different_prediction_period}. In the first two periods, the number of deaths is relatively low and stable. There is an increase in the number of deaths in the third period. Finally, we can see a huge increase or decrease in the number of deaths in the last two prediction periods. In scenarios with relatively stable trends, the County model provides better predictions than that of the Naive model. With a more drastic change in the trend, a model trained with mixup data augmentation generalizes better, hence outperforming both Naive and County models. When combining predictions from the County and Mixup models, {\ensemble} model provides a balanced and more accurate prediction across different periods.

\begin{table}[ht]
    \centering
    \begin{tabular}{>{\raggedright}p{4cm}>{\raggedleft}p{3.5cm}>{\raggedleft}p{3.5cm}>{\raggedleft\arraybackslash}p{3.5cm}}
    \hline
    Prediction Period & Model & Week 1 & Week 2\\
    \hline
    \multirow{4}{*}{2020-08-23 to 2020-09-24}
        & Naive & 26.6183 & 28.9819\\
        & County & 23.2328 & 25.2783 \\
        & Mixup & 25.3481 & 26.5330 \\
        & {\ensemble} & 23.5522 & 25.3955 \\
    \hline
    \multirow{4}{*}{2020-09-25 to 2020-10-28}
        & Naive & 27.6227 & 41.1483 \\
        & County  & 24.0228 & 31.3968 \\
        & Mixup & 27.0248 & 37.3008 \\
        & {\ensemble} & 24.1806 & 32.2430\\
    \hline
    \multirow{4}{*}{2020-10-29 to 2020-12-01}
        & Naive & 59.7121 & 95.9124 \\
        & County & 48.3641 & 62.4764 \\
        & Mixup & 51.0948 & 64.5994 \\
        & {\ensemble} & 49.1359 & 61.9947 \\
    \hline
    \multirow{4}{*}{2020-12-02 to 2021-01-17}
        & Naive  & 80.5957 & 106.7175 \\
        & County & 74.7763 & 98.2125\\
        & Mixup & 71.6854 & 84.4188\\
        & {\ensemble} & 70.8481 & 86.3737 \\
    \hline
     \multirow{2}{*}{2021-01-18 to 2021-03-14}
        & Naive  & 84.5587 & 122.9295 \\
     \multirow{2}{*}{(Use last trained model)}
        & County & 75.8700 & 82.0913\\
    
        & Mixup & 76.3645 & 80.7036\\
        & {\ensemble} & 75.0621 & 79.7467 \\
    \hline
    \end{tabular}
    \caption{Different prediction periods for the weekly total number of deaths. The prediction metrics reported is MAE.}
    \label{tab:different_prediction_period}
\end{table}

\begin{figure}[!hbt]
    \centering
        \includegraphics[width=1\textwidth, height=0.4\textwidth]{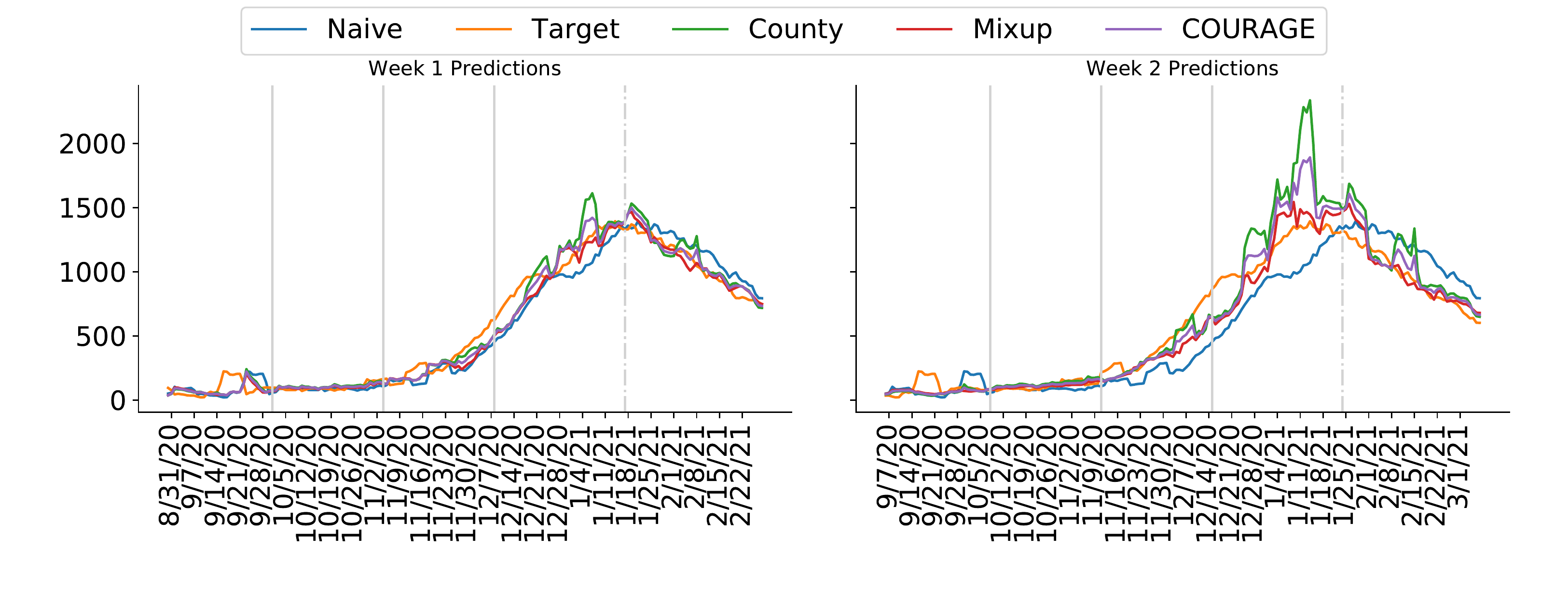}

     \caption{New York's weekly total number of deaths for Week 1 (left) predictions and Week 2 (right) predictions. Vertical lines separate different prediction periods as in Table \ref{tab:different_prediction_period}. The last dashed  vertical line marks the prediction period of recent data using our last trained model. ``Target'' is the true reported number of deaths of New York. More plots for other major states are presented in Supplementary Information.}
     \label{fig:timeline}
\end{figure}

\subsection{Comparison With Benchmark Models}
We obtain all publicly available predictions from the CDC website\footnote{https://www.cdc.gov/coronavirus/2019-ncov/covid-data/forecasting-us.html} for comparison \cite{karlen2020characterizing, Ray2020.08.19.20177493, bpagano, microsoft, oliverwyaman, cmu_delphi, harvardsimulation, lanl, umass_mb, Rodr2020, wang2020spatiotemporal,mobs, srivastava2020fast, lega2021parameter, wu2021deepgleam, uga-ceid, LSHTM, covidcomplete}. Different research groups or individuals contribute towards these predictions. CDC compiles predictions from every model that submits their weekly predictions and uploads them to the CDC website. Note that all prediction models have their strengths in different date ranges. Moreover, they outperform most of the simple methods. We compare our models' predictions with those who send predictions to the CDC official website. The comparison is done for predictions from November 11, 2020 to February 6, 2021. For each model, 
we calculate MAE for Week 1 and Week 2 predictions, and we calculate the average MAE. We summarize the result in Table \ref{tab:models_comparison1}.
Then we rank each model using the average MAE obtained for both Week 1 predictions and Week 2 predictions. Due to the extensively long list of models, we only show results from the top $10$ models in both tables. From Table \ref{tab:models_comparison1}, we can observe that our County model produces competitive accuracy for the number of deaths prediction. Our County model is a top $10$ model in terms of prediction accuracy. When we augment our dataset, our Mixup model further improves the prediction accuracy. We wish to examine which period mixup data augmentation contributes the most to model accuracy. Mixup data augmentation improves our model when a new trend emerges and helps our model to achieve better prediction in the last period, as illustrated in Table S1 of Supplementary Information. By combining strengths from the County and Mixup models, our {\ensemble} provides a good balance across different prediction periods.


\begin{table}[!ht]
    \centering
    \begin{tabular}{>{\raggedright}p{4cm}>{\raggedleft}p{3.5cm}>{\raggedleft}p{3.5cm}>{\raggedleft\arraybackslash}p{3.5cm}}
    \hline
        Model & Week 1 & Week 2 & Average\\
    \hline
       Karlen \cite{karlen2020characterizing} & 56.2163 & 77.3440 & 66.7801 \\
Ensemble \cite{Ray2020.08.19.20177493} & 57.3227 & 79.1330 & 68.2278 \\
\underline{\textbf{\ensemble}} & 61.1206 & 76.9096 & 69.0151 \\
\underline{Mixup} & 61.7890 & 77.1330 & 69.4610 \\
UMass-MB \cite{umass_mb} & 57.1578 & 84.5514 & 70.8546 \\
Oliver Wyman \cite{oliverwyaman} & 58.9645 & 85.6277 & 72.2961 \\
MOBS \cite{mobs} & 60.4486 & 84.9539 & 72.7012 \\
\underline{County} & 62.7057 & 83.5585 & 73.1321 \\
GT-DeepCOVID \cite{Rodr2020} & 67.1962 & 90.1549 & 78.6756 \\
USC \cite{srivastava2020fast} & 68.1082 & 93.3954 & 80.7518 \\
\hline
    \end{tabular}
 \caption{Comparison among different models for average MAE (from 2020-11-07 to 2021-02-06)}
    \label{tab:models_comparison1}
\end{table}

\section{Discussion}
From our results, we see that our {\ensemble} model gains its strengths from both member models, which are the County model and the Mixup model. The two core ideas associated with these member models are the aggregation of the high-quality county predictions (County model) and data augmentation (Mixup model).

The aggregation of high-quality county predictions results in high-quality state predictions. We could see high-quality county-level predictions from County and Mixup methods from Table \ref{tab:county_level_prediction}. This result shows our model and training are effective in extracting and utilizing the information from the input data. When we aggregate these accurate county-level predictions, our models can produce state-level predictions that outperform baseline models. As an illustrative comparison, the State model (baseline model) uses only state-level data as input. Such input is limited and of coarser grain, making it harder for the State model to produce accurate predictions. As a result, the State model's performance pales when compared with our model, as well as the Naive model. One way to refine state-level prediction is through the use of county-level data. By training our models using the larger county-level dataset, we obtain accurate county-level predictions. When we sum these high-quality county-level predictions, we obtain accurate state-level predictions. We justify our intuition from state-level predictions of Table \ref{tab:county_level_prediction}, by showing a strong result from our model.

The data augmentation from Mixup method also played a significant role in improving the accuracy of our predictions Due to the fast-evolving dynamics of {\covid}, most models that submit their predictions to the CDC have high accuracy only for a certain period. Existing trends are much easier to fit than emerging trends, and any changes of an existing trend are hard to predict due to the scarcity of data available at the onset of changes. The fundamental reason for prediction deterioration is the lack of visibility of the new trend given the limited data available. If we could increase the amount of available data by incorporating new data, predictive models could be improved when a new trend emerges. However, data generation is hard for any new trend. One way to improve the model's prediction on emerging trends is to improve its generalization with augmented input data. Our Mixup model uses a well-proven data augmentation technique to create new input data. Such training helps our model achieve accurate predictions in the emergence of unseen data. 

By combining the strength of each member model (the County model and the Mixup model), we obtain the {\ensemble} model. Our {\ensemble} model obtains its prediction by averaging County prediction and Mixup prediction. This simple ensemble method allows our {\ensemble} model to decrease the variance in different models and achieve a balanced model.

While our {\ensemble} model shows strong results, one limitation of our current model is that the self-attention matrix from the encoder is not easily translated to an explainable pattern. Our model's predictions use information from the confirmed cases, deaths, population, and mobility data. Their interaction is encoded by the encoder, with self attention as a key mechanism. Hence, it will be beneficial to the research community if we could see any important relationship among these inputs through the attention matrix. Besides, our model is a relatively concise model that leverages temporal data. In our future work, we plan to extend our work to include spatial information such as interaction among counties or major cities. We expect that the inclusion of such geographical information would further improving our model's predictions. Currently, our model predicts two weeks ahead predictions. We plan to include predictions of a longer horizon up to four weeks ahead predictions, and also generate a probabilistic forecast that explicitly accounts for forecast confidence.

\section{Methods}
In this section, we present our data sources and data processing used in this paper. We also present details of our transformer-based model, mixup data augmentation technique, and training procedure.
\subsection{Data Sources}
\label{sec:dataset}
We use three comprehensive datasets in this study, including confirmed cases, deaths, population, and community mobility from two sources. We only focus on states in the mainland of the United States and do not consider Hawaii, Alaska, and other unincorporated territories in this paper. We use data from $47$ states and $3206$ counties.

\textbf{Confirmed cases and deaths of Covid-19} \quad We use data from the JHU CSSE Covid-19 dataset\footnote{https://github.com/CSSEGISandData/COVID-19}\cite{pmid32087114}. This publicly available data is a curated dataset from different sources. The data used is collected from January 22, 2020, to February 7, 2021. We use confirmed cases and deaths from the dataset for every county from $47$ targeted states. 

\textbf{Community Mobility} \quad Mobility data have been shown to help the risk analysis of {\covid} and enhance the predictions of the {\covid} related deaths and cases \cite{pmid33404371, zhu2020highresolution, james2021efficiency, chicchi2021mobilitybased, pmid33622148,Carroll2020}. In current article, we use Google mobility data\footnote{https://www.google.com/covid19/mobility/} \cite{mobility} to incorporate mobility information in our model. These reports record a community's daily movement by county for different areas such as retail and recreation, groceries and pharmacies, parks, transit stations, workplaces, and residential. There is a baseline day's value for every day of a week, which represents the usual community mobility value. The baseline day's value is the median of the data from January 3, 2020, to February 6, 2020. Google mobility data reports the movement pattern of the US community.  Each day's data is in the percentage of changes from the baseline day's value. A negative value for a particular day indicates a decrease in time spent at the area from the baseline day, while a positive value indicates an increase in mobility from the baseline day.

\textbf{Demographic information}
\quad We use population information from the JHU CSSE Covid-19 dataset\cite{pmid32087114} which includes population data for each corresponding county.

\subsection{Data Preparation}
\label{sec:preprocess}
We retrieve features required for our model training, including the number of confirmed cases, the number of deaths, population, and mobility information from our datasets. We consolidate all input data into state-level and county-level datasets. We also include smoothed (average over past seven days) confirmed cases and deaths as our input features. We separate total data into training and testing datasets for each corresponding level. To test our method for a different period, we try different amounts (50\%, 60\% and 70\%, and 80\%) of the total data as our training dataset, leaving the remaining data as our testing dataset. Since we use multiple features as inputs, we apply standardization to the inputs to accommodate differences in scale for each input. We also use additional data from February 8, 2021, to March 14, 2021, to test the model trained using 80\% of the data in order to examine whether our trained model can continue to perform well for a new period without additional training.


\subsection{Transformer-based Model}
\label{sec:transformer}

\begin{figure}[htb!]
\vspace{-1em}
    \centering
    \includegraphics[width=0.6\textwidth]{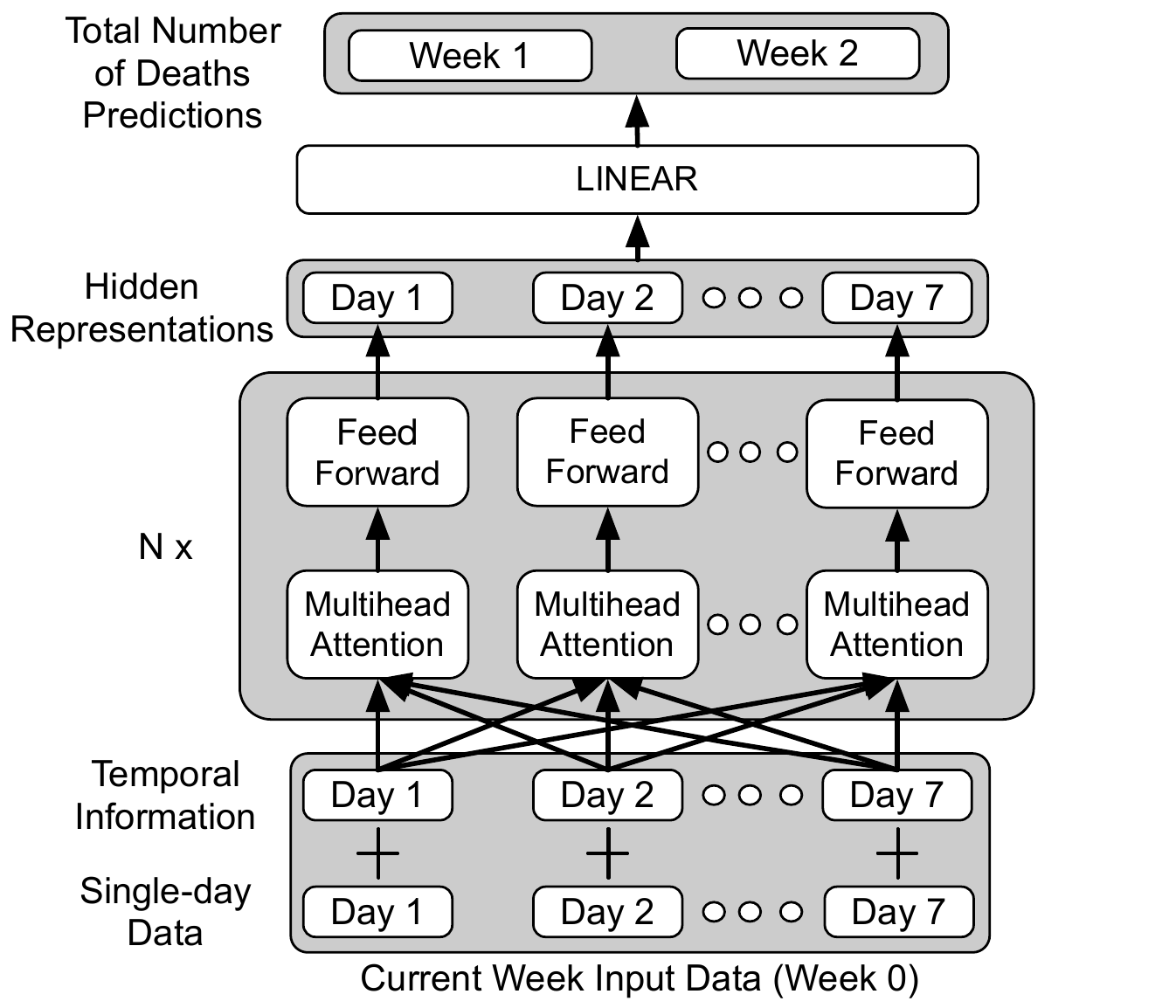}
    \caption{Transformer based model architecture}
    \label{fig:transformer}
\vspace{-0.7em}
\end{figure}

The prediction of {\covid} related number of deaths given a sequence of input is a time series modeling problem. In a typical time series prediction setting, given a sequence of previous days' number of deaths, the aim is to predict the number of deaths for the next future day. For our current article, our prediction problem is different from this typical time series problem. In the current article, our input consists of the current week's number of deaths, number of confirmed cases, smoothed (average over $7$ days) number of confirmed cases, smoothed number of deaths, community mobility data, and population of the area. More importantly, instead of predicting the daily number of deaths, our model predicts the weekly total number of deaths for the next two weeks (Week 1 and Week 2), given the current week (Week 0) input data. We join the current week input information together to form a data vector of dimension $K$. The input for our {\covid} prediction problem can be viewed as a sequence of $K$-dimensional vectors. Suppose we are given a sequence $\mathcal{S} = \{k_j\}_{j=1}^L$ of $L$ days data, where each single-day data $k_j \in \mathbf{R}^K$, occurs at time $j$.

The key ingredient of our transformer-based model is the self-attention module. Different from RNNs, the attention mechanism does not have recurrent structure. In order to incorporate the temporal information into the inputs, we use the original positional encoding method \cite{vaswani2017attention} to our data vector. Alternatively, we could also use other positional encoding methods such as relative position\cite{shaw2018selfattention} to provide the temporal information for each of single-day data vector in our input sequence.

Before passing to the attention module, we first transform our sequence of single-day data vectors using a matrix $\mathbf{U} \in \mathbf{R}^{M \times K}$. After the transformation, for any single-day data $x_j$ and its corresponding time stamp $j$, the temporal vector $z_j$ and the single-day data vector $\mathbf{U}k_j$ both reside in $\mathbf{R}^M$. Given a sequence of $L$ days data $\mathcal{S} = \{k_j\}^L_{j=1}$, we get 
\begin{equation}
    \mathbf{X} = (\mathbf{UE} + \mathbf{Z})^T,
\end{equation}
where $\mathbf{E} = [k_1,k_2,\dots,k_L] \in \mathbf{R}^{K \times L}$ is the sequence of single-day data vectors, $\mathbf{Z} = [z_1, z_2,\dots, z_j] \in \mathbf{M \times L}$ is the concatenation of the temporal vectors. 



We pass $\mathbf{X}$ through the self-attention module. Specifically, we compute the attention output $\mathbf{S}$ by
\begin{equation}
    \mathbf{S} = \text{Softmax }\left(\dfrac{\mathbf{QK}^{T}}{\sqrt{M_K}}\right)\mathbf{V},\text{ where }\mathbf{Q} = \mathbf{XW}^Q, \mathbf{K} = \mathbf{XW}^K, \mathbf{V} = \mathbf{XW}^V. 
\end{equation}
Here, $\mathbf{Q}$, $\mathbf{K}$, $\mathbf{V}$ are the query, key and value matrices obtained by different linear transformations of $\mathbf{X}$, and $\mathbf{W}^Q$, $\mathbf{W}^K \in \mathbf{R}^{M \times M_K}$, $\mathbf{W}^V \in \mathbf{R}^{M \times M_V}$ are weights for the respective linear transformations. In practice, multi-head self-attention increases model flexibility and is beneficial for data fitting. In multi-head self-attention, different sets of weights $\{\mathbf{W}^Q_h,\mathbf{W}^K_h,\mathbf{W}^V_h\}^H_{h=1}$ are used to compute different attention outputs $\mathbf{S}_1,\mathbf{S}_2,\dots,\mathbf{S}_H$. The final attention output for the sequence is then obtained by concatenating all the attention outputs and passing through the final linear transformation.
\begin{equation}
    \mathbf{S} = \left[\mathbf{S_1},\mathbf{S_2},\dots,\mathbf{S_H} \right]\mathbf{W}^O
\end{equation}
where $\mathbf{W}^O \in \mathbf{R}^{HM_V \times M}$ is an aggregation matrix.

We highlight that the self-attention mechanism allows the selection of any single-day data whose occurrence time is at any distance from the current time. The $j$-th column of the attention score from the Softmax$(\mathbf{QK}^T/\sqrt{M_K})$ indicates the extent of dependency of $j$-th single-day data ($k_j$) on its history. Hence, attention mechanism allows the capturing of short and long term dependencies of the sequence data. On the other hand, RNN-based models encode the data's history sequentially via hidden representations of events, where the state of $j$ depends on that of $j-1$, which in turn depends on $j-2$, etc. If the RNN fails to learn sufficient information for single-day data at $j$, subsequent hidden representation of any other single-day data at $t$ where $t \geq j$ will be adversely impacted. 

The attention output $\mathbf{S}$ is then fed through a position-wise feed forward neural network, generating a hidden representation $\mathbf{h}(j)$ of the input data sequence:
\begin{equation}
    \mathbf{H} = \text{ReLU}(\mathbf{SW_1^{FF}} +\mathbf{b}_1)\mathbf{W_2^{FF}} +\mathbf{b}_2, \; \mathbf{h}(j) = \mathbf{H}(j,:).
\end{equation}
Here $\mathbf{SW_1^{FF}} \in \mathbf{R}^{M \times M_H}$ , $\mathbf{SW_2^{FF}} \in \mathbf{R}^{M_H \times M}$, $\mathbf{b}_1 \in \mathbf{R}^{M_H}$,$\mathbf{b}_2 \in \mathbf{R}^{M}$ are the corresponding weights and biases of the feed forward neural networks. The resulting matrix $\mathbf{H} \in \mathbf{R}^{L \times M}$ contains hidden representations of all the information in the input sequence, where each row corresponds to a particular information. We use this final representation as an input to our linear decoder layer and obtain our predictions of the weekly total number of deaths for next two weeks.

In a typical time series prediction setting, the number of deaths prediction only forcasts the next day given the current week data. In such typical time series prediction, we need to implement masking for the attention mechanism to prevent ``peeking into the future'' issue. The masking allows any $j$-th data to attend only to any $t$-th data where $t \leq j$.  In the current article, our model is predicting the {weekly total number of deaths for the next two weeks (Week 1 and Week 2)}, given the current week (Week 0) input data. This setting frees us from such masking requirement since the model is implicitly masked from accessing the future total number of deaths from the current week data.

A transformer based model allows us to stack multiple self-attention modules together, and inputs are passed through
each of these modules sequentially. In this way our model is able to capture high level dependencies. We remark that stacking RNN/LSTM are susceptible to gradient explosion and gradient vanishing, rendering the stacked model more difficult to train. Figure \ref{fig:transformer} illustrates the architecture of our transformer-based model used in this project.

\subsection{Mixup Data Augmentation}
Data augmentation is a commonly used technique to improve the deep learning models' generalization. One recent augmentation method is mixup \cite{zhang2018mixup, guo2019augmenting, verma2019manifold} data augmentation. In mixup data augmentation, given $\mathbf{X}$ as the input space of total training data and $\mathbf{Y}$ as the corresponding output values space, each training set is a pair of $(x_i, y_i)$. Mixup {data augmentation constructs new data by interpolating it from existing data.
\begin{align*}
    \tilde{x} &= \lambda x_i + (1 - \lambda)x_j\\
    \tilde{y} &= \lambda y_i + (1 - \lambda)y_j
\end{align*}
where $(x_i, y_i)$ and $(x_j, y_j)$ are two examples drawn at random and $\lambda \in [0, 1]$. In our model, we use mixup data augmentation at input layer.
%


\subsection{Training Objective}
We train the transformer model using the Huber loss function. Specifically, the training objective is defined as 
\begin{align*}
    \underset{\theta}{\text{min}}\;\ell \left (f_{\theta}(\mathbf{X}),\mathbf{Y} \right) &= \dfrac{1}{n}\sum_{i=1}^{n} z_{i},\quad
    \text{where }z_{i} =
        \begin{cases}
        \dfrac{1}{2} (f_\theta(x_i) - y_i)^2  , & \text{if } |f_\theta(x_i) - y_i| \leq \delta \\
        \delta \; \left( |f_\theta(x_i) - y_i| - \dfrac{1}{2}  \delta \right), & \text{otherwise,}
        \end{cases}    
\end{align*}
$\mathbf{X}$ and $\mathbf{Y}$ are the input space and the target space, with a pair of testing sample as $(x_i, y_i)$. We have $n$ samples, and $\delta$ is a tuning hyperparameter.
\subsection{Training Details}
We use the transformer-based model (refer to Section \ref{sec:transformer}) for predicting both county-level and state-level number of deaths. Specifically, we use a transformer encoder of $32$ model dimensions, $1$ encoder layer with $8$ attention heads and $64$ feedforward dimensions. The output of the encoder layer connects to a single linear layer decoder for predicting both weekly total number of deaths for next two weeks (Week 1 and Week 2) using the current week input data. We use Adam \cite{kingma2017adam} optimizer for its superior empirical performance in training a neural network and set 0.001 as our initial learning rate. For every 100 epochs, we decay the learning rate by half for a total of 500 epochs of training. Figure \ref{fig:overview} illustrates our training process.

\begin{figure}[ht]
\vspace{-0.5em}
    \centering
    \includegraphics[width=0.75\textwidth]{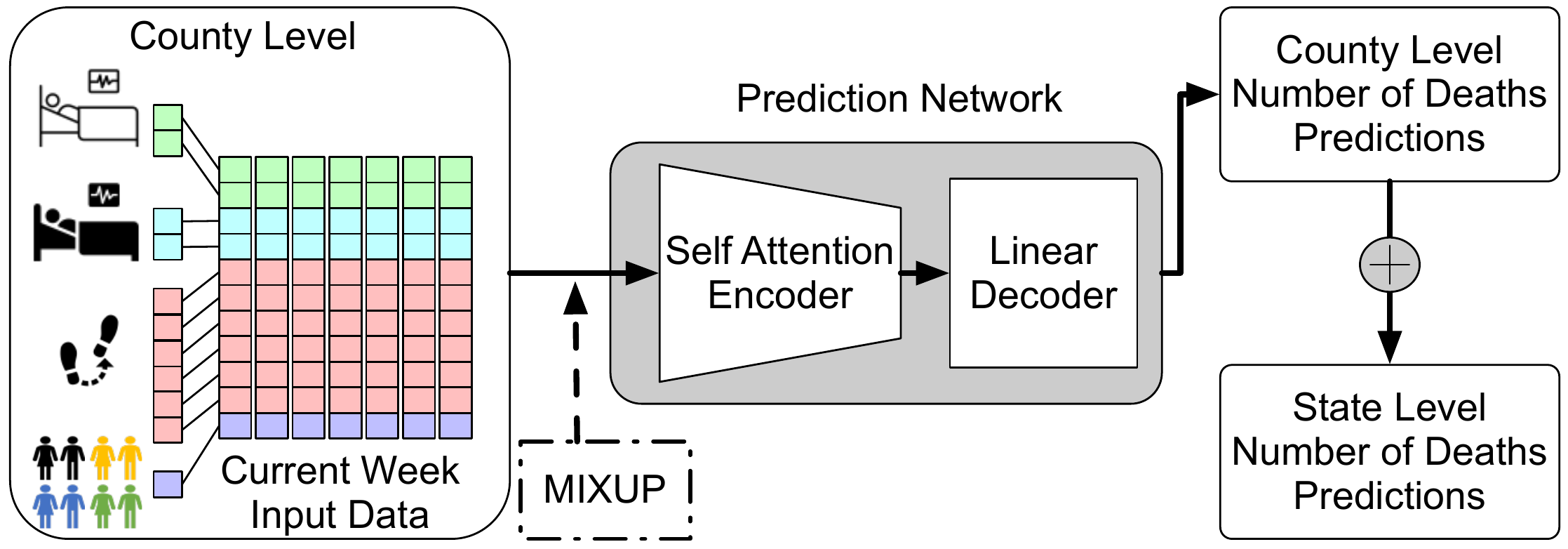}
    \caption{Overview of Prediction Flow. The county-level and state-level predictions for weekly total number of deaths are for the next week (Week 1) and the second week (Week 2) from the current week (Week 0) input data.}
    \label{fig:overview}
\end{figure}
\vspace{-0.5em}
We use both the state-level and county-level training sets to train our models. We use the smoothed number of deaths as our prediction target during training. Upon completion, all the models are used to predict weekly total number of deaths for the next two weeks (Week 1 and Week 2) using county-level dataset. We sum all the county predictions of the corresponding state as the predictions of that state. We denote the model trained as the County model. For the Mixup model, there is an additional mixup data augmentation applied to the input layer during the training phase. Our {\ensemble} model is an ensemble model of two member models, the County model and the Mixup model. {\ensemble} takes the average of both predictions from the County model and the Mixup model as its prediction. Our baseline model is a Naive model that takes the current week total number of deaths as both Week 1 and Week 2 total number of deaths predictions. For the state-level prediction task, we have another baseline model --- a State model that uses only state-level data as input when making the predictions. 


\section{Conclusion}
In summary, this article presents the new model {\ensemble} for {\covid} predictions at county level and state level for the United States. We use county-level data to train {\ensemble} and obtain state-level predictions through aggregating high quality county-level predictions. We improve our model using mixup augmentation and ensemble predictions from both the County and Mixup models as our final output. To the best of our knowledge, our model is the first model that use mixup data augmentation to improve the accuracy of {\covid} related number of deaths prediction. Our experiment shows that this new application of mixup data augmentation helps improve the model's prediction accuracy when new trends occur. {\ensemble} is a flexible model, that each member model (the County model and the Mixup model) can be used as a standalone model to produce accurate predictions in different periods. When both member models are ensembled together, {\ensemble} achieves accurate predictions across all periods.} {\covid} is a serious crisis affecting our daily life and economy. Accurate predictions of disease dynamics is a challenging task, especially when new trends emerge. We hope that through our new training method, we can improve {\covid} number of deaths predictions and provide insight for resource allocation and disease control planning.


\bibliography{ref}



\section*{Author contributions statement}

S.E., S.Y., and T.Z. designed the research; S.E., S.Y., and T.Z. performed the research; S.E. analyzed data; and S.E., S.Y., and T.Z. wrote the paper. S.Y. and T.Z. contributed equally to this work.

\subsection*{Competing interests}
The authors declare that they have no conflict of interest.

\newpage

\setcounter{page}{1}

\rfoot{\thepage}

\section*{Supplementary Information for\\
\textbf{COU}nty agg\textbf{R}egation mixup \textbf{A}u\textbf{G}m\textbf{E}ntation (COURAGE) {COVID-19} Prediction}

\subsection*{Siawpeng Er, Shihao Yang, Tuo Zhao}

\noindent Correspondence to: shihao.yang@isye.gatech.edu, tourzhao@gatech.edu

\vskip0.2in

\noindent This PDF file includes:
\begin{itemize}
\item Supplementary Table S1
\item Supplementary Figure S1
\end{itemize}

\setcounter{table}{0}
\renewcommand{\thetable}{S\arabic{table}}%
\setcounter{figure}{0}
\renewcommand{\thefigure}{S\arabic{figure}}%
\setcounter{section}{0}

\section*{Supplementary Table}
\vspace{-1em}

\begin{table}[ht]
    \centering
    \begin{tabular}{>{\raggedright}p{4cm}>{\raggedleft}p{3.5cm}>{\raggedleft}p{3.5cm}>{\raggedleft\arraybackslash}p{3.5cm}}
    \hline
        Model & Week 1 & Week 2 & Average\\
    \hline
        \underline{{Mixup} } & 70.4711 & 85.9666 & 78.2188 \\
        \underline{\textbf{\ensemble} } & 70.8875 & 87.6565 & 79.2720 \\
        Ensemble \cite{Ray2020.08.19.20177493} & 67.2675 & 93.3830 & 80.3252 \\
Karlen \cite{karlen2020characterizing} & 69.1216 & 94.2766 & 81.6991 \\
UMass-MB \cite{umass_mb} & 66.5562 & 100.0608 & 83.3085 \\
MOBS \cite{mobs}& 68.9179 & 99.7629 & 84.3404 \\
Oliver Wyman \cite{oliverwyaman} & 69.0729 & 103.6900 & 86.3815 \\
\underline{County } & 74.3374 & 98.5684 & 86.4529 \\
GT-DeepCOVID \cite{Rodr2020} & 76.1277 & 102.5137 & 89.3207 \\
Naive & 81.9271 & 107.9848 & 94.9559 \\
USC \cite{srivastava2020fast} & 82.1277 & 114.5714 & 98.3495 \\
\hline
    \end{tabular}
    \caption{Comparison among different models for average MAE (from 2020-12-19 to 2021-02-06). When new trend data is the main contribution of total prediction period, the mixup data augmentation helps to improve the model's accuracy.} 
    \label{tab:models_comparison2}
\end{table}

\section*{Supplementary Figure}
\vspace{-1em}
\begin{figure}[!ht]
    \centering
     \begin{subfigure}[b]{1\textwidth}
        \includegraphics[width=1\textwidth,height=0.35\textwidth]{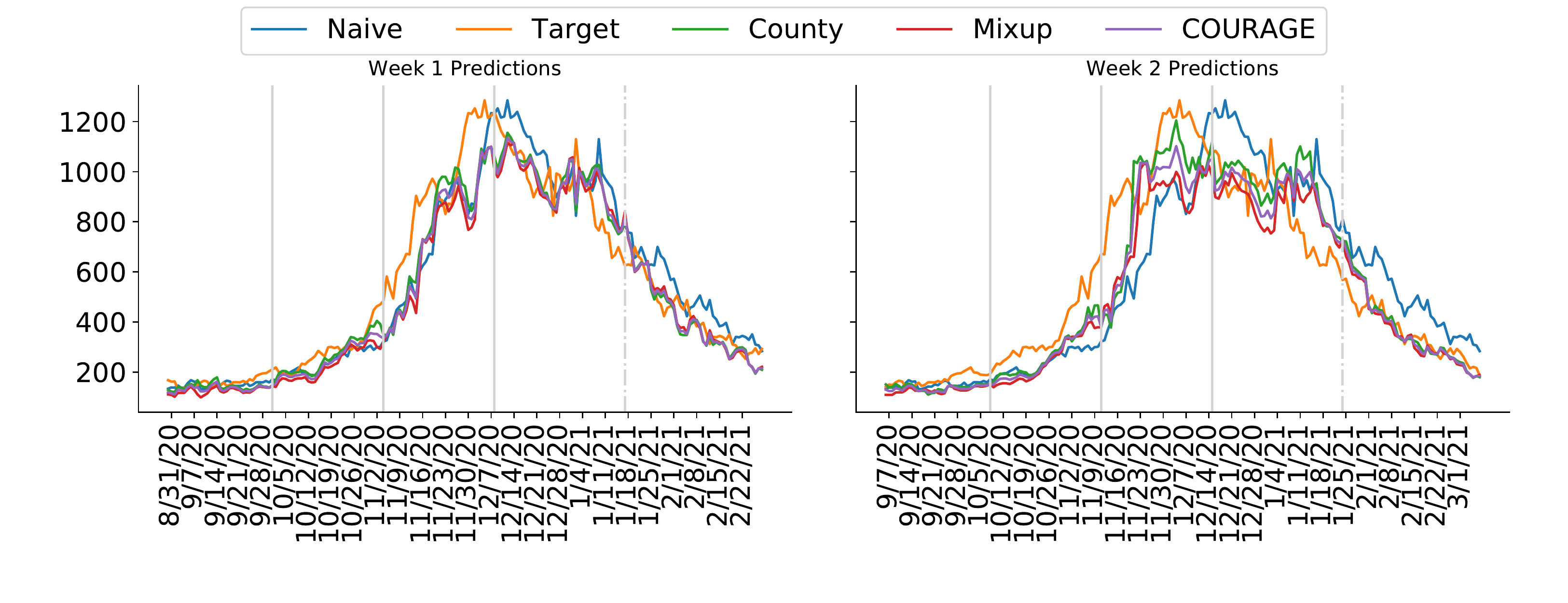}
        \vspace{-2em}
        \caption{Illinois}
    \end{subfigure}
\end{figure}
\begin{figure}[!htb]\ContinuedFloat
    \centering
      \begin{subfigure}[b]{1\textwidth}
      \includegraphics[width=1\textwidth,height=0.35\textwidth]{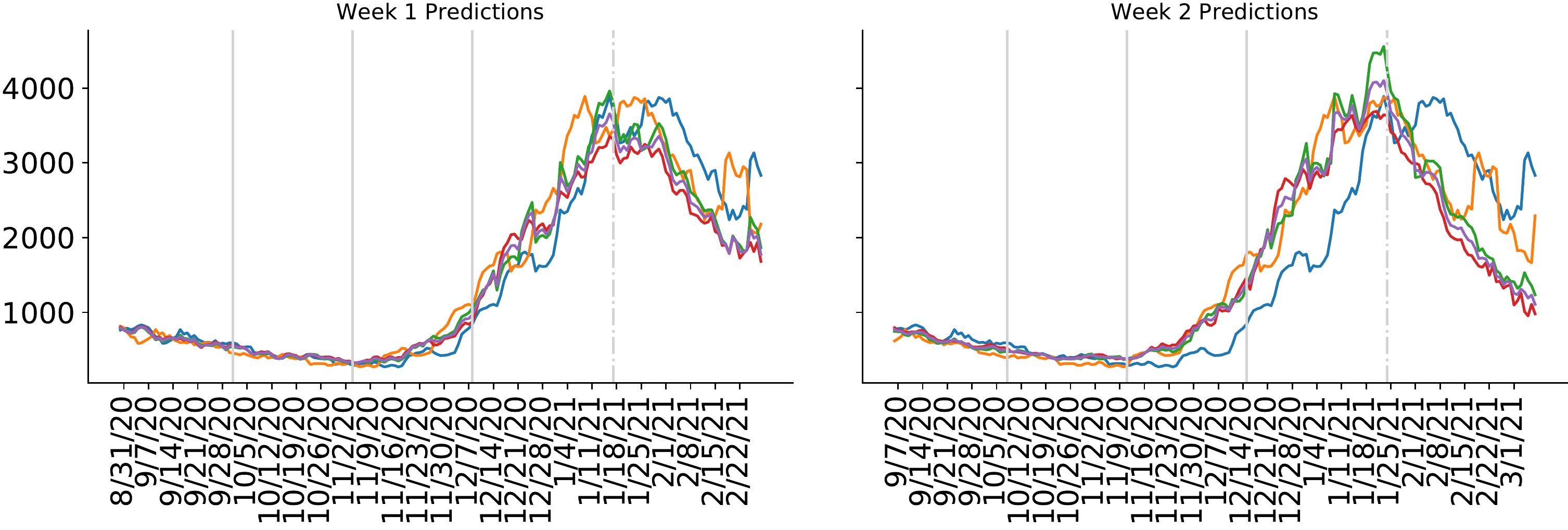}
      \vspace{-2em}
        \caption{California}
    \end{subfigure}
      \begin{subfigure}[b]{1\textwidth}
      \includegraphics[width=1\textwidth,height=0.35\textwidth]{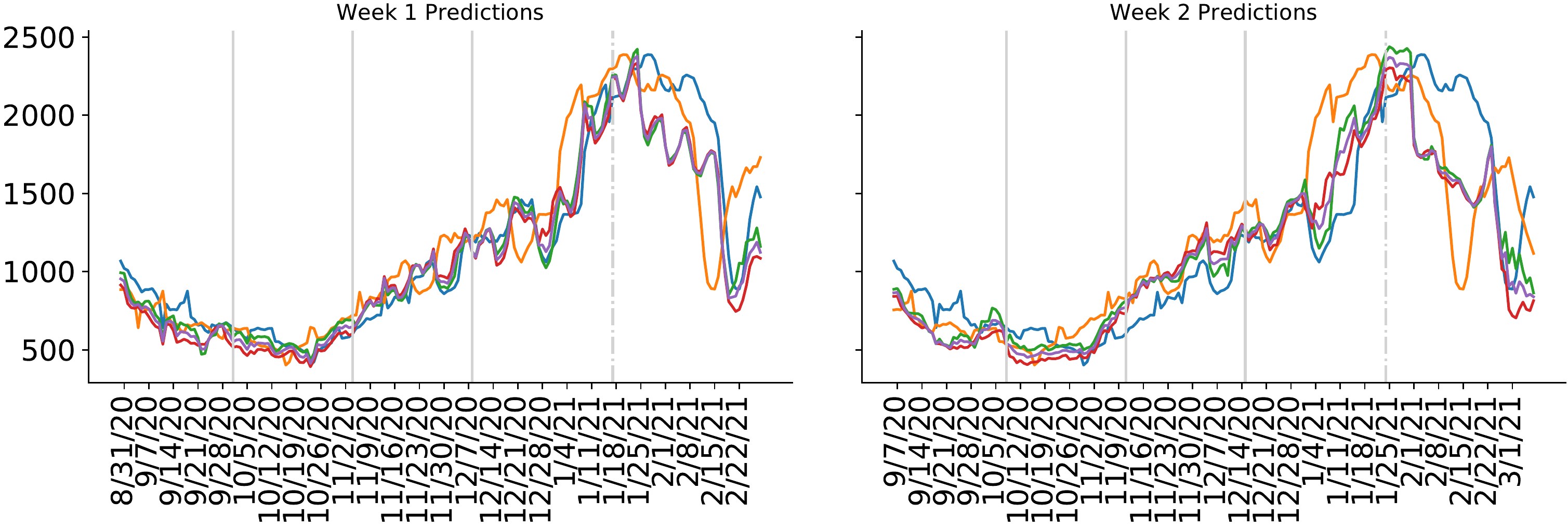}
      \vspace{-2em}
        \caption{Texas}
    \end{subfigure}

     \hfill
     \begin{subfigure}[b]{1\textwidth}
      \includegraphics[width=1\textwidth,height=0.35\textwidth]{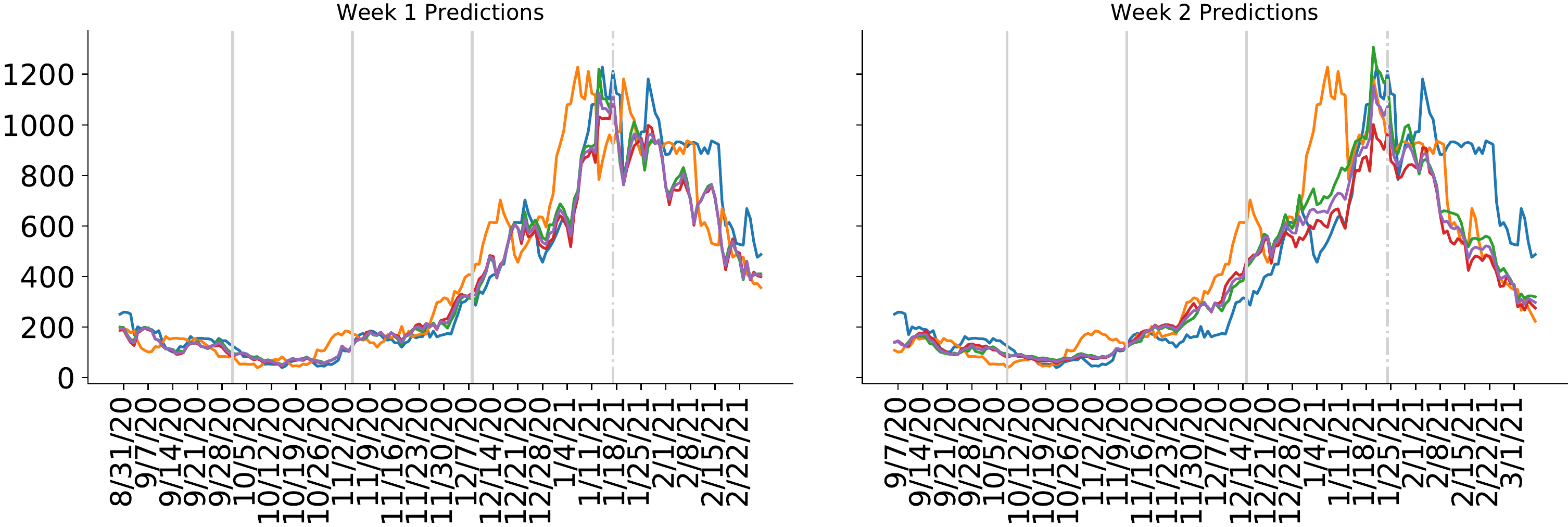}
      \vspace{-2em}
        \caption{Arizona}
    \end{subfigure}
     \caption{Weekly total number of deaths  for Week 1 (left) predictions and Week 2 (right) predictions for Illinois, California, Texas, and Arizona. Vertical lines separate different prediction periods as in Table 2 in main article. The last dashed vertical line marks the prediction period of recent data using our last trained model. ``Target'' is the true reported number of deaths for the corresponding state.} 
     \label{fig:timeline_supplementary}
\end{figure}

\end{document}